\documentclass[11pt]{article}
\usepackage{naaclhlt2013}
\usepackage{latexsym,epsfig,subfigure,pnnamed,times,url}
\usepackage{amsfonts ,amsmath}
\usepackage{lingmacros,cgloss4e}
\usepackage{tikz,tikz-dependency}
\usepackage{algpseudocode}
\usepackage{algorithmicx}

\title{Fast $k$-best Sentence Compression}

\author{Katja Filippova \& Enrique Alfonseca \\
 Google Research \\
 {\url{katjaf|ealfonseca@google.com}}}

\date{}

\begin{document}

\maketitle

\begin{abstract}
  A popular approach to sentence compression is to formulate the task
  as a constrained optimization problem and solve it with integer
  linear programming (ILP) tools. 
  Unfortunately, dependence on ILP may make the compressor
  prohibitively slow, and thus approximation techniques have been
  proposed which are often complex and offer a moderate gain in speed.
  As an alternative solution, we introduce a novel compression
  algorithm which generates $k$-best compressions relying on local
  deletion decisions.
  Our algorithm is two orders of magnitude faster than a recent
  ILP-based method while producing better compressions.
  Moreover, an extensive evaluation demonstrates that the quality of
  compressions does not degrade much as we move from single best to
  top-five results.
\end{abstract}

\section{Introduction}

There has been a surge in sentence compression research in the past
decade because of the promise it holds for extractive text
summarization and the utility it has in the age of mobile devices with
small screens.
%
Similar to text summarization, \emph{extractive} approaches which do
not introduce new words into the result have been particularly
popular.
There, the main challenge lies in knowing which words can be
deleted without negatively affecting the information content or
grammaticality of the sentence.
%
%
%
Given the complexity of the compression task (the number of possible
outputs is exponential), many systems frame it, sometimes combined
with summarization, as an ILP problem which is then solved with
off-the-shelf tools \cite{martins09,berg-kirkpatrick11,thadani13}.
While ILP formulations are clear and the translation to an ILP problem
is often natural \cite{clarke08}, they come with a high solution cost
and prohibitively long processing times \cite{woodsend12,almeida13}.
Thus, robust algorithms capable of generating informative and
grammatically correct compressions at much faster running times are
still desirable.

Towards this goal, we propose a novel supervised sentence compression
method which combines \emph{local} deletion decisions with a recursive
procedure of getting most probable compressions at every node in the
tree.
%
To generate the top-scoring compression a single tree
traversal is required. To extend the \(k\)-best list with a \(k +
1\)th compression, the algorithm needs \(O(m \times n)\) comparisons
where \(n\) is the node count and \(m\) is the average branching
factor in the tree.
Importantly, approximate search techniques like beam search
\cite{galanis10,wang13}, are not required.
 
Compared with a recent ILP method \cite{filippova.emnlp13}, our
algorithm is two orders of magnitude faster while producing shorter
compressions of equal quality.
Both methods are supervised and use the same training data and
features. The results indicate that good readability and
informativeness, as perceived by human raters, can be achieved without
impairing algorithm efficiency. Furthermore, both scores remain high
as one moves from the top result to the top five.
To our knowledge we are the first to report evaluation results beyond
single best output.

To address cases where local decisions may be insufficient, we present
an extension to the algorithm where we tradeoff the guarantee of
obtaining the top scoring solution for the benefit of scoring a node
subset as a whole. This extension only moderately affects the running
time while eliminating a source of suboptimal compressions.

\paragraph*{Comparison to related work}

Many compression systems have been introduced since the very first
approaches by \newcite{grefenstette98}, \newcite{jing00a} and
\newcite{knight00}. Almost all of them make use of syntactic
information (e.g., \newcite{clarke06b,mcdonald06a,toutanova07}), and
our system is not an exception. Like \newcite{nomoto09,wang13} we
operate on syntactic trees provided by a state-of-the-art parser.
The benefit of modifying a given syntactic structure is that the space
of possible compressions is significantly constrained: instead of all
possible token subsequences, the search space is restricted to all the
subtrees of the input parse. While some methods \emph{rewrite} the
source tree and produce an alternative derivation at every consituent
\cite{knight00,galley07}, others \emph{prune} edges in the source tree
\cite{filippova.inlg08,galanis10,wang13}. Most of these approaches are
supervised in that they learn from a parallel compression corpus
either the rewrite operations, or deletion decisions.
In our work we also adopt the pruning approach and use parallel data
to estimate the probability of deleting an edge given context.

Several text-to-text generation systems use ILP as an optimization
tool to generate new sentences by combining pieces from the input
\cite{clarke08,martins09,woodsend10b,filippova.emnlp13}. While
off-the-shelf general purpose LP solvers are designed to be fast, in
practice they may make the compressor prohibitively slow, in
particular if compression is done jointly with summarization
\cite{berg-kirkpatrick11,qian13,thadani14}.
Recent improvements to the ILP-based methods have been significant but
not dramatic. For example, \newcite{thadani14} presents an
approximation technique resulting in a 60\% reduction in average
inference time.
Compared with this work, the main practical advantage of our system is
that it is very fast without trading compression quality for speed
improvements. On the modeling side, it demonstrates that local
decisions are sufficient to produce an informative and grammatically correct
sentence.

Our recursive procedure of generating \(k\) best compressions at every
node is partly inspired by frame semantics \cite{fillmore03} and its
extension from predicates to any node type \cite{titov11}.
The core idea is that there are two components to a high-quality
compression at every node in the tree: (1) it should keep all the
essential arguments of that node; (2) these arguments should
themselves be good compressions. This motivates an algorithm with a
recursively defined scoring function which allows us to obtain
\emph{k}-best compressions nearly as fast as the single best one.
In this respect our algorithm is similar to the \emph{k}-best parsing
algorithm by \newcite{huang05}.

\section{The Top-down Approach}\label{sec:appr}

\begin{figure*}[htp]
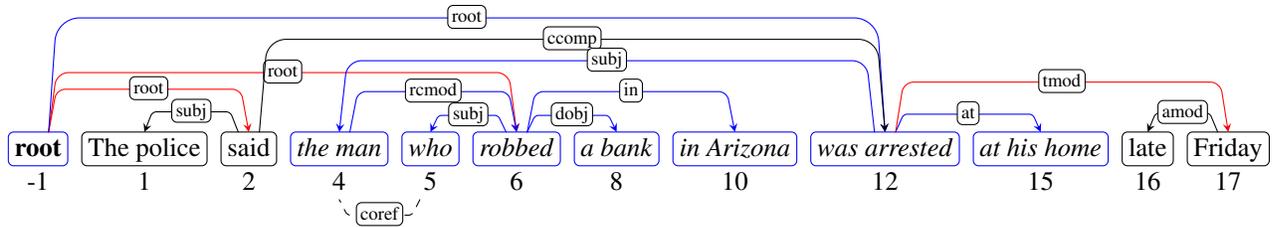

  \small
  \centering
    \begin{dependency}
      \begin{deptext}[column sep=0.2cm, row sep=.1ex]
        \textbf{root} \& The police \& said \& \emph{the man} \& \emph{who} \&
        \emph{robbed} \& \emph{a bank} \& \emph{in Arizona} \& \emph{was arrested} \& \emph{at his home}
        \& late \& Friday \\
        -1 \& 1 \& 2 \& 4 \& 5 \& 6 \& 8 \& 10 \& 12 \& 15 \& 16 \& 17 \\
      \end{deptext}
      \depedge[edge unit distance=1.7ex]{3}{2}{subj}
      \depedge[edge style=red, edge unit distance=1.8ex]{1}{3}{root}
      \depedge[edge style=red, edge unit distance=1.0ex]{1}{6}{root}
      \depedge[edge style=blue, edge unit distance=1.2ex]{1}{9}{root}
      \depedge[edge unit distance=1.3ex]{3}{9}{ccomp}

      \depedge[edge style=blue, edge unit distance=1.7ex]{4}{6}{rcmod}
      \depedge[edge style=blue, edge unit distance=1.4ex]{6}{5}{subj}
      \depedge[edge style=blue, edge unit distance=1.4ex]{6}{7}{dobj}
      \depedge[edge style=blue, edge unit distance=1.7ex]{6}{8}{in}

      \depedge[edge style=blue, edge unit distance=1.2ex]{9}{4}{subj}
      \depedge[edge style=blue, edge unit distance=1.5ex]{9}{10}{at}
      \depedge[edge style=red, edge unit distance=1.4ex]{9}{12}{tmod}
      \depedge[edge unit distance=1.6ex]{12}{11}{amod}

      \depedge[edge unit distance=1.3ex, edge below, arrows=-, style={dashed}]{5}{4}{coref}

      \wordgroup[group style=blue]{1}{1}{1}{r}
      \wordgroup{1}{2}{2}{p}
      \wordgroup{1}{3}{3}{s}
      \wordgroup[group style=blue]{1}{4}{4}{m}
      \wordgroup[group style=blue]{1}{5}{5}{w}
      \wordgroup[group style=blue]{1}{6}{6}{r}
      \wordgroup[group style=blue]{1}{7}{7}{b}
      \wordgroup[group style=blue]{1}{8}{8}{z}
      \wordgroup[group style=blue]{1}{9}{9}{a}
      \wordgroup[group style=blue]{1}{10}{10}{h}
      \wordgroup{1}{11}{11}{l}
      \wordgroup{1}{12}{12}{f}
    \end{dependency}
    \vspace*{-.9cm}
  \caption{Transformed parse tree of the example sentence. The
    compression subtree is highlighted with blue color.}
  \label{fig:tree}
\end{figure*}
 
Our approach is syntax-driven and operates on dependency trees (Sec.\
\ref{sec:appr:prep}). The input tree is pruned to obtain a valid
subtree from which a compression is read off.
The pruning decisions are carried out based on predictions of a
maximum entropy classifier which is trained on a parallel corpora
with a rich feature set (Sec.\ \ref{sec:appr:feat}).
Section \ref{sec:appr:best} explains how to generate the single,
top-scoring compression; Section \ref{sec:appr:k} extends the idea to
arbitrary \(k\).

\subsection{Preprocessing}\label{sec:appr:prep}

Similar to previous work, we have a special treatment for function
words like determiners, prepositions, auxiliary verbs.
Unsurprisingly, dealing with function words is much easier than
deciding whether a content word can be removed. Approaches which use a
constituency parser and prune edges pointing to constituents, deal
with function words implicitly \cite{berg-kirkpatrick11,wang13}. 
Approaches which use a dependency representation either formulate hard
constraints \cite{almeida13}, or collapse function words with their
heads. We use the latter approach and transform every input tree
\cite{nivre06} following \newcite{filippova.inlg08} and also add edges
from the dummy root to finite verbs. Finally, we run an entity tagger
and collapse nodes referring to entities.

Figure \ref{fig:tree} provides an example of a transformed tree with
extra edges from the dummy root node and an undirected coreference
edge for the following sentence to which we will refer throughout this
section:

\enumsentence{\label{ex:sent} %
  The police said the man who robbed a bank in Arizona was arrested at his home late Friday.  }

\subsection{Estimating deletion probabilities}\label{sec:appr:feat}

The supervised component of our system is a binary maximum entropy
classifier \cite{berger96} which is trained to estimate the
probability of \emph{deleting} an edge given its local context. 
In what follows, we are going to refer to two probabilities,
$p_{del}(e)$ and $p_{ret}(e)$:

\vspace{-.5cm}
\begin{equation}
  p_{del}(e_{n,m}) + p_{ret}(e_{n,m}) = 1,
\end{equation}
\noindent
where \emph{del} stands for \emph{deleting}, \emph{ret} stands for
\emph{retaining} edge \emph{e} from node \emph{n} to node \emph{m},
and $p_{del}(e_{n,m})$ is estimated with MaxEnt.

The features we use are inspired by most recent work
\cite{almeida13,filippova.emnlp13,wang13} and are as follows:
\begin{description}
\item[syntactic:] edge labels for the child and its siblings;
  NE type and PoS tags;
\item[lexical:] head and child lemmas; negation; concatenation of
  parent lemmas and labels;
\item[numeric:] depth; node length in words and characters; children
  count for the parent and the child.
\end{description}
\noindent
Note that no feature refers to the compression generated so far and
therefore the probability of removing an edge needs to be calculated
only once on a first tree traversal.

Assuming that we have a training set comprising pairs of a transformed
tree, like the one in Figure \ref{fig:tree}, and a compression subtree
(e.g., the subtree covering all the nodes from \emph{the man} to
\emph{at his home}), the compression subtrees provide all the negative
items for training (blue edges in Fig.~\ref{fig:tree}). The positive
items are all other edges originating from the nodes in the
compression (red edges). The remaining edges (black) cannot be used for
training.

Although we chose to implement hard constraints for function words
(see Sec.\ \ref{sec:appr:prep} above), we could also apply no tree
transformations and instead expect the classifier to learn that,
e.g., the probability of deleting an edge pointing to a determiner is
zero. However, given the universality of these rules, it made more
sense to us to encode them as preprocessing transformations.

\subsection{Obtaining top-scoring compression}\label{sec:appr:best}

To find the best compression of the sentence we start at the dummy
root node and select a child \(n\) with the highest \(p_{ret}(e_{root,
  n})\). The root of the example tree in Figure \ref{fig:tree} has
three children (\emph{said$_2$, robbed$_6$, was
  arrested$_{12}$}). Assuming that $p_{ret}$'s for the three predicates are
\(.07, .5, .9\), the third child is selected.
From there, we recursively continue in a top-down manner and
at every node \(n\) whose children are \(M = \{m_1, m_2, ...\}\)
search for a children subset \(C_n \subseteq M\) maximizing

\vspace{-.5cm}
\begin{equation}\label{eq:score:cn} 
\begin{split}
  score(C_n) = & \sum_{m \in M \setminus C_n} \log p_{del}(e_{n, m})
  \\
  & + \sum_{m \in C_n} \log p_{ret}(e_{n, m}).
\end{split}
\end{equation} 

\noindent
Since $p_{del}$ and $p_{ret}$ sum to one, this implies that every
edge with $p_{ret} < 0.5$ is deleted.
However, we can take any \(\rho \in [0,1]\) to be a threshold for deciding
between keeping vs.\ deleting an edge and linearly scale  \(p_{del}\) and \(p_{ret}\) so that after
scaling \(\hat{p}_{del} < 0.5\) if and only if \(p_{del} < \rho\).
Of course, finding a single \(\rho\) value that would be universally
optimal is hardly possible and we will return to this point in
Sec.~\ref{sec:ext}.

Consider the node \emph{was arrested} in Figure \ref{fig:tree} and its
three children listed in Table \ref{table:arrested} with \(p_{ret}\)
given in brackets. 

\begin{table}[htb]
  \begin{center}
    {\small
      \begin{tabular}{c | c | c}
        & \emph{was arrested}\(_{12}\) &  \\
        \hline
        \emph{the man\(_{4}\)} (1.0) &\emph{at his home\(_{15}\)} (.22)
        & \emph{Friday\(_{17}\)} (.05) \\
      \end{tabular}
    }
  \end{center}
  \caption{Arguments of \emph{was arrested} with their $p_{ret}$'s.}
  \label{table:arrested}
\end{table}
\noindent
With \(\rho = 0.5\), the top scoring subset is \(C_{12} = \{4\}\), its
score being \(0 + \log .78 + \log .95\).  The next step is to decide
whether node 4 (\emph{the man}) should retain its relative clause
modifier or not. There is no need to go further down the \emph{Friday}
node and consider the score of its sole argument (\emph{late}).

\subsection{From top-scoring to \(k\)-best}\label{sec:appr:k}

A single best compression may appear too long or too short, or fail to
satisfy some other requirement. In many cases it is desirable to have
a pool of \emph{k-best} results to choose from and in this subsection
we will present our algorithm for efficiently generating a
\emph{k}-best list (summarized in Fig.~\ref{fig:algo}).

First, let us slightly modify the notation used up to this point to be
able to refer to the \emph{k}th best result at node \emph{n}. Instead
of \(C_n \subseteq M\), we are going to use \(C^k_n\), where \(k \in
\mathbb N \cup \{-1\}\). Unlike \(C_n\), every \(C^k_n\) is an ordered
sequence of exactly \(|M|\) elements, corresponding to \emph{n}'s
children:

\begin{equation}
  C^k_n = [C^{k_1} _{m_1}, C^{k_2}_{m_2}, ..., C^{k_{|M|}}_{m_{|M|}}].
\end{equation}

\noindent
For every child $m_i$ not retained in the compression, the superscript
$k_i$ is \emph{-1}. For example, for the singleton subset $C_{12}$
containing only node 4 in the previous subsection the corresponding
best result \(C^0_{12}\) is:

\begin{equation}\label{eq:ex:best}
  C^0_{12} = [C^0_4, C^{-1}_{15}, C^{-1}_{17}]. 
\end{equation}
\noindent
Note that at this point we do not need to know what $C^0_4$ actually
is. We simply state that the best result for node 12 must include the
best result for node 4.

The scoring function for \(C^k_n\) is the averaged sum of the scores
of \emph{n}'s chlidren and must be decreasing over $k \geq 0$
($score(C^{k+a}_n) \leq score(C^k_n), a >0$):

\vspace{-.5cm}
\begin{equation}\label{eq:score:ckn}
  score(C^k_n) = \frac{1}{|M|} \sum_{C^{k_i}_{m_i} \in C^k_n} score(C^{k_i}_{m_i}).
\end{equation}

\noindent
When \(k \in \{-1, 0\}\), i.e., when we either delete a child or take
its best compression, the score is the familiar probabilities:
\begin{equation}\label{eq:score:ckn:base}
  score(C^{k}_m) =
  \begin{cases}
    \log p_{del}(e_{n, m}) & \text{if } k = -1 \\
    \log p_{ret}(e_{n, m}) &  \text{if } k = 0 \\
  \end{cases}
\end{equation}

\noindent
Greater values of \emph{k} correspond to \emph{k+1}'th best result at
node \emph{n}. Consider again node 12 from Table
\ref{table:arrested}. The \emph{k}-best results at that node may
include any of the following variants (the list is not complete):
\begin{center}
\vspace{-.5cm}
  {\small
  \begin{tabular}{l l l}
  \([C^0_4, C^{-1}_{15}, C^{-1}_{17}],\) & \([C^0_4, C^0_{15},
  C^{-1}_{17}],\) & 
  \([C^2_4, C^{-1}_{15}, C^0_{17}],\)\\
  \( [C^1_4, C^{-1}_{15}, C^{1}_{17}],\) & \([C^0_4, C^0_{15}, C^{1}_{17}]\)
  & \([C^{-1}_4, C^{0}_{15}, C^{0}_{17}]\).
 \end{tabular}
}
\end{center}

How should these be scored so that high quality compressions are
ranked higher? Our assumption is that the quality of a compression at
any node is subject to the following two conditions:
\begin{enumerate}
\item The child subset includes essential arguments and does not
  include those that can be omitted.
\item The variants for the children retained in the
  compression are themselves high-quality compressions.
\end{enumerate}
\noindent
For example, a compression at node 12 which deletes the first child
(\emph{the man}) is of a poor quality because it misses the subject
and thus violates the first condition. A compression which retains the
first node but with a misleading compression, like \emph{the man robbed
  in Arizona} (\(C^k_4 = [C^l_6], C^l_6 = [C^{-1}_5, C^{-1}_8,
C^0_{10}]\)), is not good either because it violates the second
condition, which is in turn due to the first condition being violated in $C^l_6$.
Hence, a robust scoring function should balance these two
considerations and promote variants with good compression at every
node retained. 
Note that for finding the single best result it is sufficient to focus
on the first condition only, ignoring the second one, because the best
possible result is returned for every child, and the scoring function
in Eq.~\ref{eq:score:cn} does exactly that. However, once we begin to
generate more than a single best result, we start including
compressions which may no longer be optimal. So the main challenge in
extending the scoring function lies in how to propagate the
scores from node's descendants so that both conditions are satisfied.

Given the best result at node \emph{n}, which is obtained in a single
pass (Sec.~\ref{sec:appr:best}), the second best result must be one of
the following:
\begin{itemize}
\item The next best scoring child subset whose score we know how
  compute from Eq.~(\ref{eq:score:ckn}-\ref{eq:score:ckn:base}) (e.g.,
  for node 12 it would be \([C^0_4, C^0_{15}, C^{-1}_{17}]\), see
  Eq.~\ref{eq:ex:best}).
\item A subset of the same children as the best one but with one of
  \(k_i\)'s which were \emph{0} in the best result increased to
  \emph{1} (e.g., for node 12 it would be \([C^1_4, C^{-1}_{15},
  C^{-1}_{17}]\), see Eq.~\ref{eq:ex:best}):
\end{itemize}
\noindent
No other variant can have a higher score than either of these. Unless
there is a tie in the scores, there is a single new second-best
subset. And it follows from the decreasing property and the definition
of the scoring function that if more than a single \(k_i\) is
increased from zero, the score is lower than when only one of the
\(k_i\)'s is modified. For example, \(score([C^2_4, C^{-1}_{15},
C^1_{17}]) \leq score([C^0_4, C^{-1}_{15}, C^0_{17}]) \leq
score([C^0_4, C^0_{15}, C^{-1}_{17}])\), the latter comparison is
between two new subsets whose scores can be computed directly from
Eq.~(\ref{eq:score:ckn}-\ref{eq:score:ckn:base}).
Hence, the second best result \(C^1_n\) is either the next best
subset, or one of the at most \(|M|\) candidates. 

Assuming that \(k_j = 0\) in the best result, the score of candidate
\(C^{k^*_j}_n\) generated from \(C^0_n\) by incrementing \(k_j\) is
defined as
\vspace{-.5cm}
\begin{equation}\label{eq:score:2nd}
    score(C^{k^*_j}_n) = score(C^0_n) +
    \frac{score(C^{0+1}_{m_j})}{|M|}.
\end{equation}
\noindent
Generalizing to an arbitrary \emph{k}, the \emph{k+1}'th result is
also either an unseen subset, whose score is defined in
Eq.~\ref{eq:score:ckn}, or it can be obtained by increasing a \(k_i\)
from a non-zero value in one of the \emph{k}-best results generated so
far. Given a \(C^k_n\), the score of a candidate generated by
incrementing the value of \(k_j\) is:

\vspace{-.5cm}
\begin{equation}\label{eq:score:next}
  {\small
    \begin{split}
      score(C^{k^*_j}_n) = score(C^k_n)
      + \frac{1}{|M|} score(C^{k_j+1}_{m_j})  \\
      -
      \begin{cases}
        0 &
        \text{if } k_j = 0, \\
        \frac{1}{|M|} score(C^{k_j}_{m_j}) &
        \text{if } k_j > 0. \\
      \end{cases}
    \end{split}
  }
\end{equation}
 
Notice the similarity between Eq.~\ref{eq:score:2nd} and
Eq.~\ref{eq:score:next}. The difference is that when we explore
candidates of \(k_j\)'s greater than zero, we replace the contribution
of \(m_j\)'th child: the \(k_j\)'th best score is replaced with \(k_j
+ 1\)'th best score. However, the edge score (\(C^0_{m_j}\)) is never
taken out of the total score of \(C^k_n\). This is motivated by the
first of the two conditions above. As an illustratation to this point,
consider the predicate from Table \ref{table:arrested} one more time
and assume that \(p_{ret}(e_{17, 16}) = 0.4\), i.e., the probability
of \emph{late} being the argument of \emph{Friday} is 0.4. The
information that the temporal modifier (node 17) is an argument with a
very low score should not disappear from the subsequent scorings of
node 12's candidates. Otherwise a subsequent result may get a higher
score than the best one, violating the decreasing property of the
scoring function, as the final line below shows:

\begin{center}
  {\small
  \begin{tabular}{l l}
  \([C^0_4, C^{-1}_{15}, C^{-1}_{17}]\) & \((0 + \log .78 + \log .95)
  / 3\) \\
  \([C^0_4, C^{-1}_{15}, C^0_{17}]\) & \((0 + \log .78 + \log .05) / 3\) \\
  \([C^0_4, C^{-1}_{15}, C^1_{17}]\) & \((0 + \log .78 + \log .05 +
  \log .4) / 3\) \\
  \([C^0_4, C^{-1}_{15}, C^1_{17}]^*\) & \((0 + \log .78 + \log .4) / 3.\) 
 \end{tabular}
}
\end{center}

To sum up, we have defined a monotonically decreasing scoring function
and outlined our algorithm for generating \emph{k}-best compressions
(see Fig.~\ref{fig:algo}). As at every request the pool of candidates
for node \emph{n} is extended by not more than $|M| + 1$ candidates,
the complexity of the algorithm is $O(k \times N \times m)$ (\emph{k}
times node count times the average branching factor).

\begin{figure}
  \begin{footnotesize}
    \begin{algorithmic} 
      \Function{KBestCompress}{G, k}
      \State{\(C^0_r = \) \Call{FindBestCompression}{G}}
      \State{\emph{result } \(\gets \{C^0_r\}\), \emph{heaps } \( \gets
        \{\}\)}
      \Comment{Heaps for every node \emph{n}.}
      \While{\(|result| < k\)}
      \State{\emph{result } \(\gets\) \emph{result } \(\cup \)
        \{\Call{FindNextBest}{$C^{|result - 1|}_r$, \emph{heaps}}\}}
      \EndWhile
      \State{\Return \emph{result}}
      \EndFunction

      \State{}

      \Function{FindNextBest}{$C^k_n$, \emph{heaps}}
      \State{\Comment{Generate candidates by increasing a \(k_i\)
          in the recent result.}}
      \ForAll{\(C^{k_i}_{m_i} \in C^k_n\)}
      \If{\(k_i > -1\)}
      \Comment{Copy the result and update one field.}
      \State{\(C^{k^*_i}_n \gets C^k_n\)}
      \State{$C^{k_i + 1}_{m_i} \gets
        $\Call{FindNextBest}{$C^{k_i}_{m_i}$, \emph{heaps}} }
      \State{\(C^{k^*_i}_n[m_i] \gets C^{k_i +
          1}_{m_i}\), \Call{UpdateScore}{$C^{k^*_i}_n$}}
      \State{\emph{heaps[n]} \(\gets\) \emph{heaps[n]} \(\cup \{C^{k^*_i}_n\}\)}
      \EndIf
      \EndFor
      \State{\(C^{k^*}_n \gets \Call{GenerateNextBestSubset}{G, n}\) }
      \State{\emph{heaps[n]} \(\gets\) \emph{heaps[n]} \(\cup \{C^{k^*}_n\}\)}
      \State{\Return{\emph{ pop(heaps[n])}}} \Comment{\(C^{k+1}_n\),
        the best candidate for \emph{n}.}
      \EndFunction
      \State{}

      \Function{FindBestCompression}{G}
      \State{\(C^0_r \gets [], best \gets -1, max \gets -1\)}
      \ForAll{\(n \in children(root(G))\)}
      \State{\(C^0_r \gets C^0_r + \) \Call{FindBestResult}{n, G}}
      \If{\( p_{ret}(e_{r, n}) > max \) }
      \State{\(max \gets p_{ret}(e_{r, n}), best \gets n\)}
      \EndIf
      \EndFor
      \ForAll{\(n \in children(root(G))\)}
      \If{\(n \neq best\)} \(C^0_r[n] \gets C^{-1}_n\) \EndIf
      \EndFor
      \State{\Return \(C^0_r\)}
      \Comment{In the list, only one child is selected.}
      \EndFunction

      \State{}

      \Function{FindBestResult}{n, G} 
      \State{\(C^0_n \gets []\)}
      \ForAll{\(m \in children(n)\)}
      \If{\( p_{ret}(e_{n, m}) \geq 0.5 \) }
      \State{\(C^0_n \gets C^0_n + \) \Call{FindBestResult}{m, G}}
      \Else{ \(C^0_n \gets C^0_n + C^{-1}_m\)}
      \EndIf
      \EndFor
      \State{\Return \(C^0_n\)}
      \EndFunction

    \end{algorithmic}
    \caption{Pseudocode of the algorithm for finding \emph{k}-best
      compressions of graph \emph{G}. Obvious checks for termination
      conditions and empty outputs are not included for
      readability. $C^k_n[m]$ refers to the result for child \emph{m}
      of \emph{n} in $C^k_n$. Scoring is defined in Equations
      \ref{eq:score:ckn}-\ref{eq:score:next}. Similar to
      \newcite{huang05} we use a heap for efficiency; \emph{heaps[n]}
      refers to the heap of candidates for node \emph{n}. }
    \label{fig:algo}
  \end{footnotesize}
\end{figure}

\section{Adding a Node Subset Scorer}\label{sec:ext}

On the first pass, the top-down compressor attempts to find the best possible children subset of every node by considering every child separately and making the retain-or-delete decisions independently of one another. How conservative or aggressive the algorithm is, is determined by a single parameter \(\rho \in [0, 1]\) which places a boundary between the two decisions. With smaller values of \(\rho\) a low probability of deletion (\(p_{del}\)) would suffice for a node child to be removed. Conversely, a greater value of \(\rho\) would mean that only children about which the classifier is fairly certain that they must be deleted would be removed. 

Unsurprisingly, the value of \(\rho\) is hard to optimize as it may be too low or too high, depending on a node. While retaining a child which could be dropped would not result in an ungrammatical sentence, omitting an important argument may make the compression incomprehensible. When doing an error analysis on a development set, we did not encounter many cases where the compression was clearly ungrammatical due to a wrongly omitted argument. However, results like that do have a high cost and thus need to be addressed.
Consider the following example:

\enumsentence{\label{ex:ablaze} %
  Yesterday the world was ablaze with the news that the CEO will step down.  }

\noindent
In this sentence, \emph{ablaze} is analyzed as an adverbial modifier of the verb \emph{to be} and the classifier assigns a score of 0.35 to the edge pointing to \emph{ablaze}. With a decision boundary above 0.35, the meaningful part of the predicate is deleted and the compression becomes incomplete. With the boundary at 0.5, the top scoring subset is a singleton containing only the subject. However, there are hardly any cases where the verb \emph{to be} has a single argument, and our algorithm could benefit from this knowledge. 

In the extended model, the score of a children subset (Eq.\ \ref{eq:score:ckn}) gets an additional summand, $\log p(|C^k_n|)$, where $|C^k_n|$ refers to the number of \emph{n}'s children actually retained in $C^k_n$, i.e., with $k_i \geq 0$:

\vspace{-.5cm}
\begin{equation}\label{eq:score:ext}
  score^*(C^k_n) = \log p(|C^k_n|) + \frac{1}{|M|} \sum_{C^{k_i}_{m_i} \in C^k_n} score(C^{k_i}_{m_i}).
\end{equation}
\noindent
Unfortunately, with the updated formula, we can no longer generate \emph{k}-best compressions as efficiently as before. However, we can keep a beam of \emph{b} subset candidates for every node and select the one maximizing the new score.

To estimate the probability of a children subset size after compression, $p(|C^k_n|)$, we use an averaged perceptron implementation \cite{freund99} and the features described in Sec.~\ref{sec:appr:feat}. We do not differentiate between sizes greater than four and have five classes in total (\emph{0, 1, 2, 3, 4+}).

\section{Evaluation}

The purpose of the evaluation is to validate the following two hypotheses, when
comparing the new algorithm with a competitive ILP-based sentence
compressor \cite{filippova.emnlp13}:
\begin{enumerate}
\item The top-down algorithm was designed to perform local decisions at each
  node in the parse tree, as compared to the global optimization carried out by
  the ILP-based compressor. We want to verify whether the local model can attain
  similar accuracy levels or even outperform the global model, and do
  so not only for the single best but the top \emph{k} results.
\item Automatic ILP optimization can be quite slow when the number of candidates
  that need to be evaluated for any given input is large. We want to quantify
  the speed-up that can be attained without a loss in accuracy by taking
  simpler, local decisions in the input parse tree.
\end{enumerate}

\subsection{Evaluation settings}

\paragraph*{Training, development and test set}

The aligned sentences and compressions were collected using the
procedure described in \newcite{filippova.emnlp13}.
%
The training set comprises 1,800,000 items, each item consisting of
two elements: the first sentence in a news article and an extractive
compression obtained by matching content words from the sentence with
those from the headline (see \newcite{filippova.emnlp13} for the
technical details). A part of this set was held out for classifiers
evaluation and development.
For testing, we use the dataset released by
\newcite{filippova.emnlp13}\footnote{http://storage.googleapis.com/sentencecomp/compression-data.json}. This
test set contains 10,000 items, each of which includes the original
sentence and the extractive compression and the URL of the source
document. From this set, we used the first 1,000 items only, leaving
the remaining 9,000 items unseen, reserved for possible future
experiments.
We made sure that our training set does not include any of the
sentences from the test set.



The training set provided us with roughly 16 million edges for
training MaxEnt with 40\% of positive examples (deleted edges). For
training the perceptron classifier we had about 6 million nodes at our
disposal with the instances distributed over the five classes as follows:

\begin{center}
  \begin{tabular}{ccccc}
    0 & 1 & 2 & 3 & 4+ \\
    \hline
    19.5\% & 40.6\% & 31.2\% & 7.9\% & 1\% 
  \end{tabular}
\end{center}


\paragraph*{Baseline}
We used the recent ILP-based algorithm of \newcite{filippova.emnlp13}
as a baseline. We trained the compressor with all the same features as
our model (Sec.~\ref{sec:appr:feat}) on the same training data using
an averaged perceptron \cite{collins02b}. To make this system
comparable to ours, when training the model, we did not provide the
ILP decoder with the oracle compression length so that the model
learned to produce compressions in the absense of length
argument. Thus, both methods accept the same input and are comparable.

\subsection{Automatic evaluation}

\begin{figure}[t]
  \vspace{-1.2cm}
  \includegraphics[width=\columnwidth]{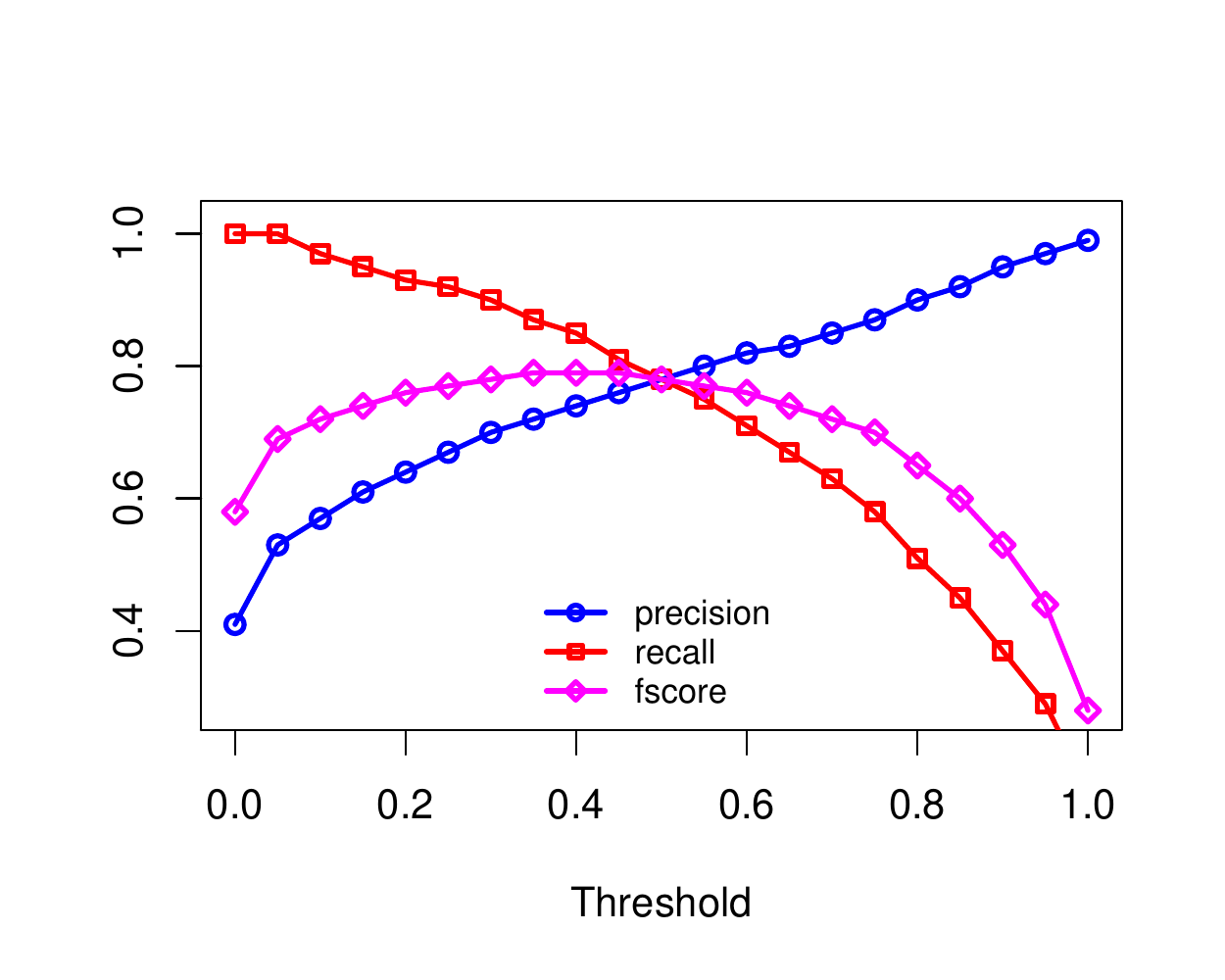}
  \vspace{-.7cm}
  \caption{Per-edge precision, recall and F1-score using different thresholds on
    the prediction values of MaxEnt.}
  \label{fig:seti-precision-recall}
\end{figure}

To measure the quality of the two classifiers (MaxEnt from
Sec.~\ref{sec:appr:feat} and perceptron from Sec.~\ref{sec:ext}), we
performed a first, direct evaluation of each of them on a small held
out portion of the training set. The MaxEnt classifier predicts the
probability of deleting an edge and outputs a score between zero and
one. Figure~\ref{fig:seti-precision-recall} plots precision, recall
and F1-score at different threshold values. The highest F1-score is
obtained at 0.45. Regarding the perceptron classifier that predicts
the number of children that we should retain for each node, its
accuracy and per-class precision and recall values are given in
Table~\ref{table:results:abe}.

\begin{table}[htb]
  \begin{center}
    \begin{small}
    \begin{tabular}{l|ccccc}
      Acc & 0 & 1 & 2 & 3 & 4+ \\
      \hline
      72.7 & 69 / 63 & 75 / 78 & 76 / 81 & 60 / 42 & 44 / 16 
    \end{tabular}
    \end{small}
  \end{center}
\caption{Accuracy and precision / recall for the classifier predicting
  the optimal children subset size.}\label{table:results:abe}
\end{table}

For an automatic evaluation of the quality of the sentence
compressions, we followed the same approach as
\cite{riezler03,filippova.emnlp13} and measured F1-score by comparing
the trees of the generated compressions to the golden, extractive
compression. Table~\ref{table:results:fscore} shows the results of the
ILP baseline and the two variants of the Top-down approach on the test
data (\textsc{Top-down + NSS} is the extended variant described in
Sec.~\ref{sec:ext}).
The NSS version, which incorporates a prediction on the number of
children to keep for each node, is slightly better than the original
Top-down approach, but the results are not statistically significant.

\begin{table}[htb]
  \begin{center}
    \begin{tabular}{l|rr}
      & \emph{F1-score} & \emph{Compr.\ rate} \\
      \hline
      \textsc{ILP}                      & 73.9 & 46.5\% \\
      \textsc{Top-down}           & 76.7 & 38.3\% \\
      \textsc{Top-down + NSS} &  77.2 & 38.1\% 
    \end{tabular}
  \end{center}
  \caption{Results for the baseline and our two algorithms.}
  \label{table:results:fscore}
\end{table}

It is important to point out the difference in compression rates
between ILP and {\sc Top-down}: 47\% vs. 38\% (the average compression
rate on the test set is 40.5\%). Despite a significant advantage due
to compression rate \cite[see next subsection]{napoles11a}, ILP
performs slightly worse than the proposed methods.

Finally, Table~\ref{table:results:topn} shows the results when
computing the F1-score for each of the top-5 compressions as generated
by the Top-down algorithms. As can be seen, in both cases there is a
sharp drop between the top two compressions but further scores are
very close. Since the test set only contains a single oracle
compression for every sentence, to understand how big the gap in
quality really is, we need an evaluation with human raters.

\begin{table}[htb]
  \begin{center}
    \begin{tabular}{c|c}
      \textsc{Top-down} & \textsc{Top-down + NSS} \\
      \hline
      76.7; 60.4; 62; 60.9; 59.6 & 77.2; 60.5; 64; 62.6; 60
    \end{tabular}
  \end{center}
  \caption{F1 scores for the top five compressions ($k = 1, 2, 3, 4, 5$).}
  \label{table:results:topn}
\end{table}

\begin{figure*}[htb]
\vspace{-1cm}
  \centerline{
    (a) \includegraphics[width=5cm]{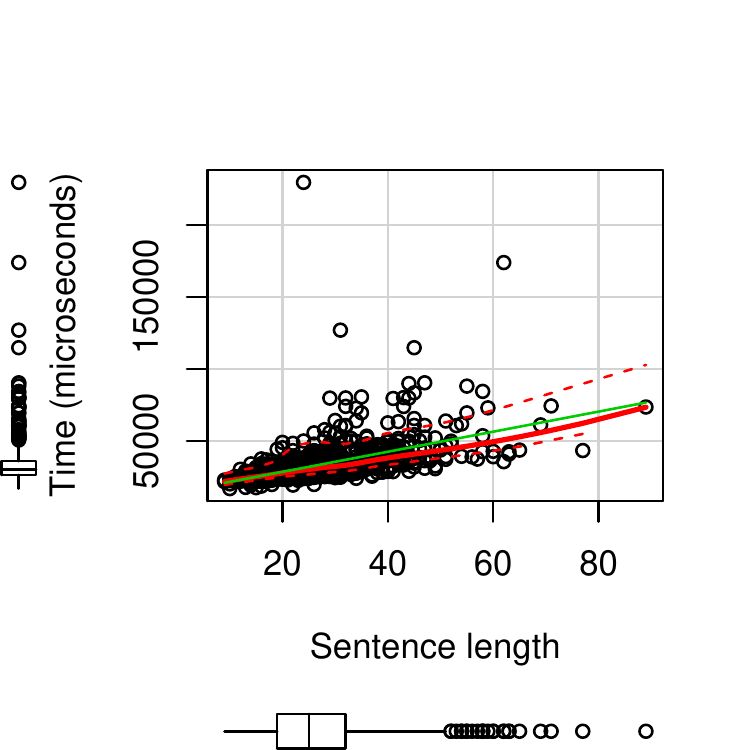}
    (b) \includegraphics[width=5cm]{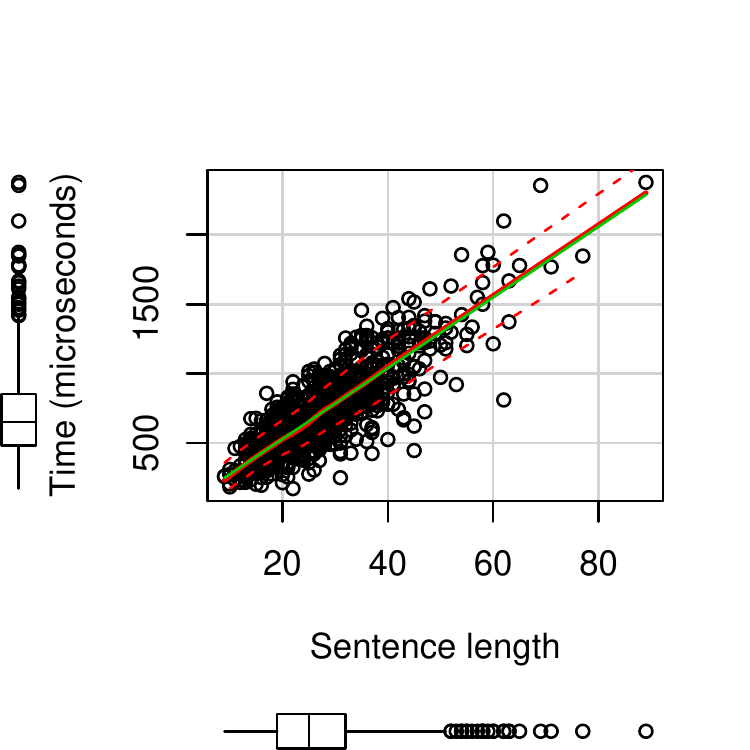}
    (c) \includegraphics[width=5cm]{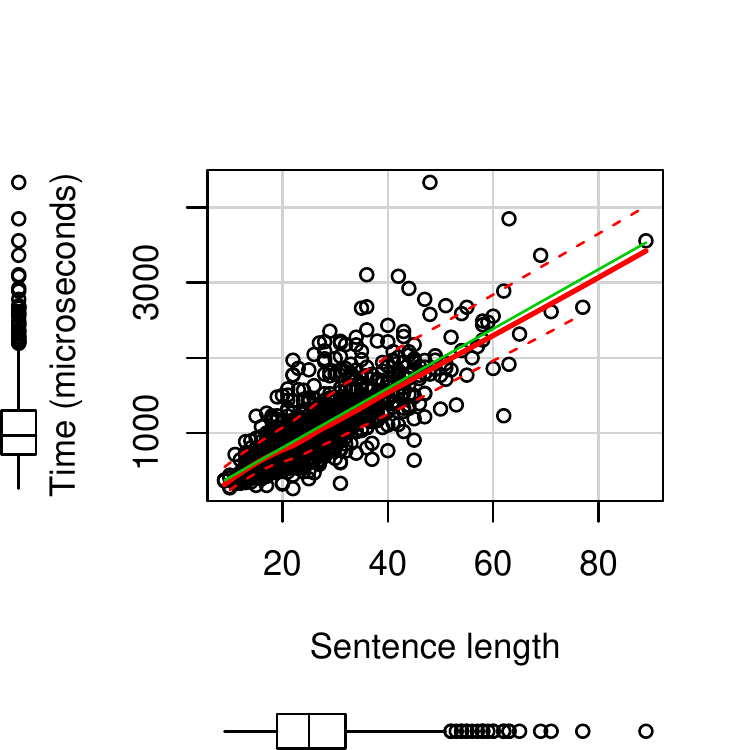}
  }
  \caption{Per-sentence processing time for the test set: (a) ILP; (b) Top-down;
    (c) Top-down + NSS.}
  \label{fig:time}
\end{figure*}
\begin{figure*}[htb]
\vspace{-1cm}
  \centerline{
    (a) \includegraphics[width=5cm]{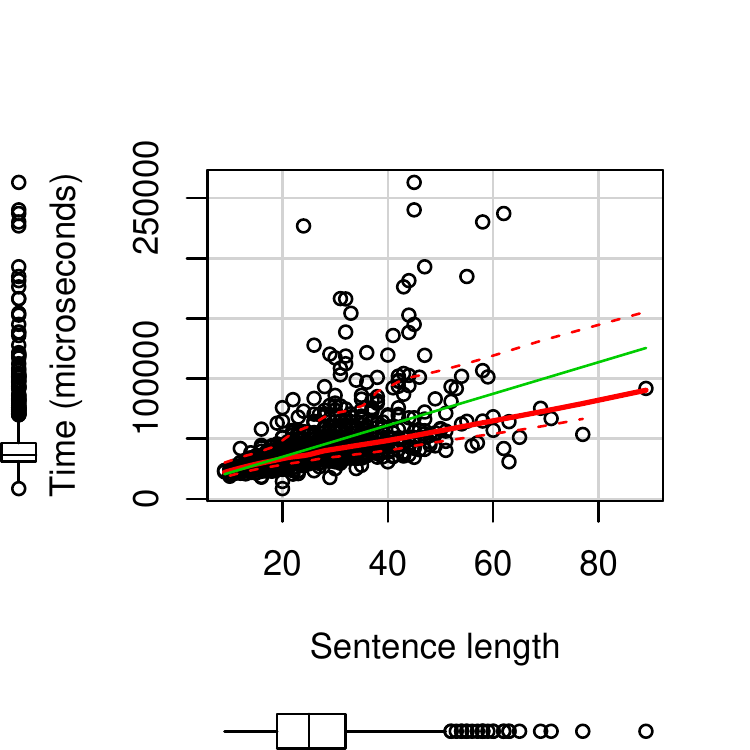}
    (b) \includegraphics[width=5cm]{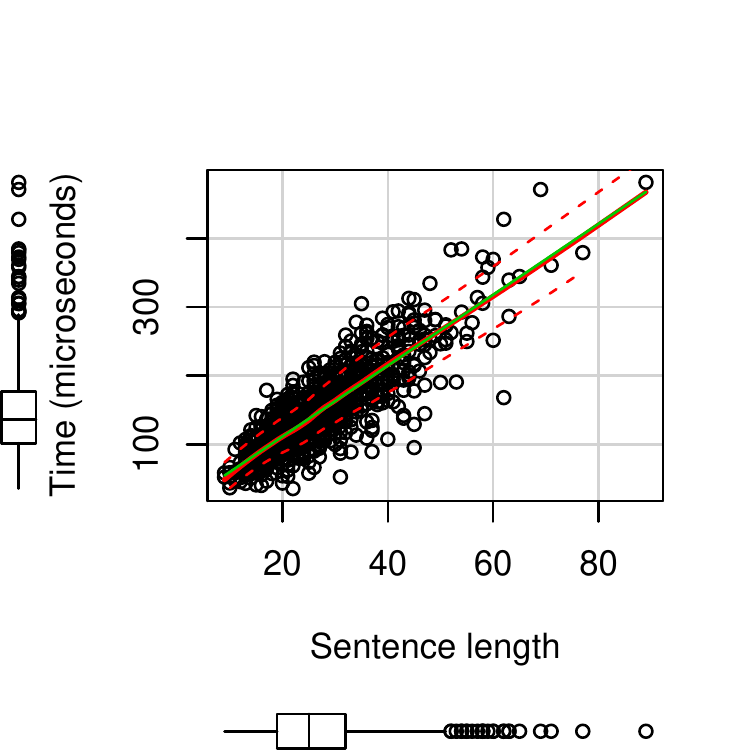}
    (c) \includegraphics[width=5cm]{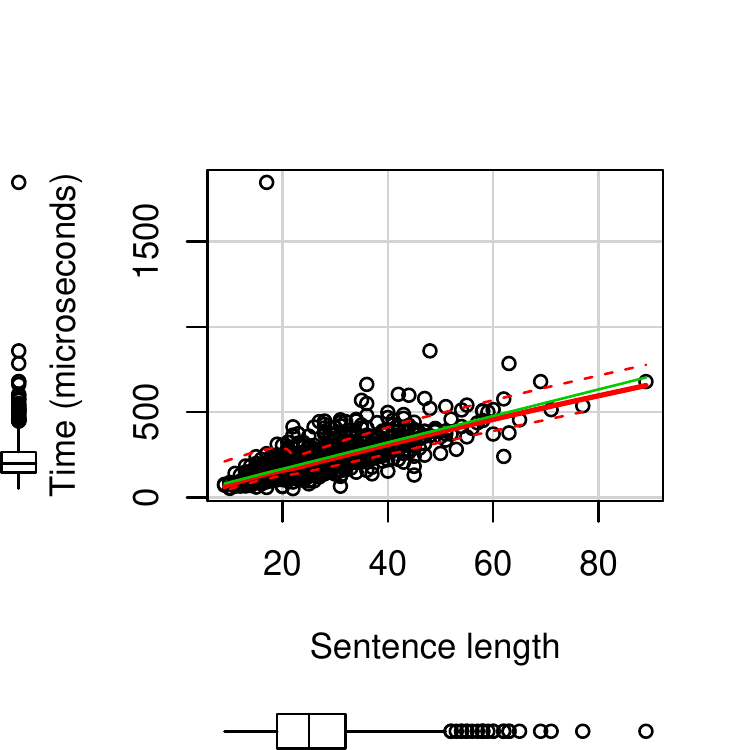}
  }
  \caption{Average processing time for getting all of the top-5
    results: (a) ILP; (b) Top-down; (c) Top-down + NSS.}
  \label{fig:time:top5}
\end{figure*}

\vspace{-.5cm}
\subsection{Manual evaluation}

The first 100 items in the test data were manually rated by humans. We
asked raters to rate both readability and informativeness of the
compressions for the golden output, the baseline and our
systems\footnote{The evaluation template and rated sentences are
  included in the supplementary material.}. For both metrics a 5-point
Likert scale was used, and three ratings were collected for every
item.
Note that in a human evaluation between \textsc{ILP} and
\textsc{Top-down (+ NSS)} the baseline has an advantage because (1) it
prunes less aggressively and thus has more chances of producing a
grammaticaly correct and informative outputs, and (2) it gets a hint
to the optimal compression length in edges.
 We have used Intra-Class Correlation (ICC)
\cite{shrout79,cicchetti94} as a measure of inter-judge agreement. ICC
for readability was 0.59 (95\% confidence interval [0.56, 0.62]) and
for informativeness it was 0.51 (95\% confidence interval [0.48,
0.54]), indicating fair reliability in both cases.

Results are shown in Tables~\ref{table:results:manual}
and~\ref{table:results:manualtopn}. As in the automatic evaluations, the two
Top-down systems produced indistinguishable results, but both are significantly
better than the ILP baseline at 95\% confidence. The top-down results are also
now indistinguishable from the extractive compressions.

\begin{table}[htb]
  \begin{center}
    \begin{tabular}{l|ll}
       & {\small Readability} & {\small Informativeness} \\
      \hline
      \textsc{Extractive}          & 4.33  & 3.84  \\
      \hline
      \textsc{ILP}                & 4.20  & 3.78 \\
      \textsc{Top-down}           & 4.41 & 3.91  \\
      \textsc{Top-down + NSS}     & 4.38 & 3.87  \\
    \end{tabular}
  \end{center}
  \caption{Results of the manual evaluation.}

  \label{table:results:manual}
\end{table}

\begin{table}[htb]
  \begin{center}
    \begin{tabular}{l|ccc}
\emph{k} & \textsc{ILP} & \textsc{Top-down} & \textsc{Top-down + NSS}  \\
      \hline
      1 & 4.20 / 3.78 & 4.41 / 3.91 & 4.38 / 3.87 \\
      2 & 3.85 / 3.09 & 4.11 / 3.31 & 4.26$^\dagger$ / 3.37 \\
      3 & 3.53 / 2.73 & 4.03$^\dagger$ / 3.37$^\dagger$ & 3.97$^\dagger$ / 3.40$^\dagger$ \\
      4 & 3.31 / 2.27 & 3.80$^\dagger$ / 3.16$^\dagger$ & 3.90$^\dagger$ / 3.19$^\dagger$ \\
      5 & 3.00 / 2.42 & 3.90$^\dagger$ / 3.41$^\dagger$ & 4.12$^\dagger$ / 3.41$^\dagger$ 
    \end{tabular}
  \end{center}
  \caption{Readability and informativeness for the top five compressions; $^\dagger$ indicates that one of the systems is statistically
    significantly better than ILP at 95\% confidence using a t-test.}
  \label{table:results:manualtopn}
\end{table}

\subsection{Efficiency}

The average per-sentence processing time is 32,074 microseconds (Intel
Xeon machine with 2.67 GHz CPU) using \textsc{ILP}, 929 using
\textsc{Top-down + NSS}, and 678 using \textsc{Top-down}. This means
that we have obtained almost a 50x performance increase over
\textsc{ILP}.
Figure~\ref{fig:time} shows the processing time for each of the 1,000
sentences in the test set with sentence length measured in tokens.

For obtaining \emph{k}-best solutions, the decrease in time is even
more remarkable: the average time for generating each of the top-5
compressions using \textsc{ILP} is 42,213 microseconds, greater than
that of the single best result. Conversely, the average time for each
of the top-5 results decreases to 143 microseconds using
\textsc{Top-down}, and 195 microseconds using \textsc{Top-down + NSS},
which means a 300x improvement. The reason is that the Top-down
methods, in order to produce the top-ranked compression, have already
computed all the per-edge predictions (and the per-node NSS
predictions in the case of \textsc{Top-down + NSS}), and generating
the next best solutions is cheap.

\section{Conclusions}

We presented a fast and accurate supervised algorithm for generating
\emph{k}-best compressions of a sentence. Compared with a competitive
ILP-based system, our method is 50x faster in generating the best
result and 300x faster for subsequent \emph{k}-best
compressions. Quality-wise it is better both in terms of readability
and informativeness. Moreover, an evaluation with human raters
demonstrates that the quality of the output remains high for the top-5
results.
%


\bibliographystyle{pnnamedshort}
\bibliography{lit}
\end{document}